\documentclass{llncs} % use larger type; default would be 10pt

\usepackage[utf8]{inputenc} % set input encoding (not needed with XeLaTeX)

\usepackage{geometry} % to change the page dimensions
\usepackage{graphicx} % support the \includegraphics command and options

\usepackage{booktabs} % for much better looking tables
\usepackage{array} % for better arrays (eg matrices) in maths
\usepackage{paralist} % very flexible & customisable lists (eg. enumerate/itemize, etc.)
\usepackage{verbatim} % adds environment for commenting out blocks of text & for better verbatim
\usepackage{framed}
\usepackage{makecell}
\usepackage{multirow}
\usepackage{subcaption}

\usepackage{fancyhdr} % This should be set AFTER setting up the page geometry
\usepackage{sectsty}
\allsectionsfont{\sffamily\mdseries\upshape} % (See the fntguide.pdf for font help)
\usepackage[nottoc,notlof,notlot]{tocbibind} % Put the bibliography in the ToC
\usepackage[titles,subfigure]{tocloft} % Alter the style of the Table of Contents

\title{Conformal Prediction and Trustworthy AI}

\author{
Anthony Bellotti\inst{1}%\orcidID{0000-0001-6317-5877} 
\and 
Xindi Zhao\inst{1}%\orcidID{0009-0009-9957-201X}
}

\institute{School of Computer Science, University of Nottingham Ningbo China
\email{anthony-graham.bellotti@nottingham.edu.cn}
}

\date{} 

\begin{document}
\maketitle

\begin{abstract}
Conformal predictors are machine learning algorithms developed in the 1990's by Gammerman, Vovk, and their research team, to provide set predictions with guaranteed confidence level.
Over recent years, they have grown in popularity and have become a mainstream methodology for uncertainty quantification in the machine learning community.
From its beginning, there was an understanding that they enable reliable machine learning with well-calibrated uncertainty quantification.
This makes them extremely beneficial for developing trustworthy AI, a topic that has also risen in interest over the past few years, in both the AI community and society more widely.
In this article
\footnote{Preprint for an essay to be published in \emph{The Importance of Being Learnable
(Enhancing the Learnability and Reliability of Machine Learning Algorithms)}
Essays Dedicated to Alexander Gammerman on His 80th Birthday, LNCS Springer Nature Switzerland AG \emph{ed.} Nguyen K.A. and Luo Z.}, 
we review the potential for conformal prediction to contribute to trustworthy AI beyond its marginal validity property, addressing problems such as generalization risk and AI governance.
Experiments and examples are also provided to demonstrate its use as a well-calibrated predictor and for bias identification and mitigation.
\end{abstract}

\section{Introduction}

Over the past ten years the increasing impact of AI in society has been extraordinary, and the impact is set to continue to increase. Although much of the excitement around AI is with generative models such as ChatGPT for text and DALL-E for images and video, many of the AI workhorses in industry and government are predictive machine learning models. 
Predictive machine learning can be used for decision-support or agentic tasks. Such supervised machine learning models can provide either a prediction or propose some action.
They are widely used in all areas of society \cite{QIAN2024100040}. For example, they can be used to predict the type of cancer from gene expression data extracted from blood sample \cite{zhang2023machine}, or used to screen loan applicants based on characteristics of an applicant such as income, outstanding debt, and employment status; e.g. \cite{GUNNARSSON2021292}.
However, with the rise in the use of AI, there has been a growing rise of concern and suspicion with the use of AI. People, governments and companies wonder if AI can be trusted to perform the tasks they are set to do \cite{10.1145/3555803}. How do we know that they are helping, and not hindering? Human medical doctors have been trained over many years, so why should we trust a machine learning model that has been trained in just a few hours?
The risks associated with using AI are broad, but most can be summarized within these seven categories:
\begin{description}
\item{\textbf{Poor performance}}. 
This is the most basic risk, that the AI does not perform as well as it should to perform reliably.
For example, in a classification problem, it would be typical to ensure that overall misclassification rate is low across an independent test dataset. For regression problems, it would be typical to use a measure such as mean-squared-error. 
Many alternative measures are available and the choice of performance measure depends on the task.
For predictive machine learning projects, this type of evaluation is standard \cite{provost2013data}. 
However, there remain some subtle problems that are sometimes overlooked: (i) Generalizability: is the model robust when presented with new data which may have shifted from the original training data in some way? This depends on both the task and the model; (ii) Calibration: how far can we trust each individual prediction? In particular, if the model provides an estimate of uncertainty, how well does the estimate match measured uncertainty?
%(iiii) lack of robustness: how good is the model beyond the examples of data given in its training?
\item{\textbf{Bias}}. 
Does the model give unequal predictions or performance across different subpopulations of data? This can give rise to discrimination. For example, in their audit of commercial face recognition systems, \cite{buolamwini2018gender} found discrimination against black women. In particular, darker-skinned females had much higher misclassification rates (up to 34.7\%) relative to lighter-skinned males (up to 0.8\%). If the subpopulations are protected classes such as gender, race or religion, then such models that exhibit bias are not only unethical, but could be illegal if deployed.
\item{\textbf{Human/AI interaction}}. 
The relationship between humans and AI is a concern. It may be that humans will become over-reliant on AI, or alternatively, humans may irrationally under-rate the value of AI. We may wonder what the extensive use of AI by humans will have on human behaviour and happiness. As humans we may be concerned that the AI ought to be able to explain itself or its decisions (transparency).
We are also concerned that AI is not \emph{misused} by humans to cause harm. For example, criminals could use AI for fraud using DeepFake technology, or it could be used to generate disinformation to influence people politically.
\item{\textbf{Objective misalignment}}. 
Although AI is largely autonomous, humans still need to specify the task they perform to the AI system. For machine learning, typically this is done using an objective function, translating the task description into an algebraic form. This can create difficulty since not all nuances of the task may be correctly and completely specified in the required precise mathematical form.
There is a risk that the objective function is misaligned with what is really required, hence exhibiting unintended consequences.
For example, \cite{amodei2016faulty} describes how they used reinforcement learning to build a system to play CoastRunner, a boat racing video game, with the objective of maximizing the score. However, instead of learning to complete the race as intended, the AI learned to drive the boat in a circle to pick up points from peripheral tasks. In a sense, the system was doing what we tasked it to do, but it was not what was \emph{intended}. This experiment was in a game setting, but in a real-world setting such as automated vehicles such behaviour could be dangerous \cite{amodei2016concrete}.
\item{\textbf{Security and Privacy}}. 
Machine learning uses data for training. There is a risk that private data is used without consent, or confidential data may be compromised during the training process, or confidential data is able to be reconstructed from the output of the algorithm. These raise serious security and privacy concerns when using AI.
\item{\textbf{Ethics, Legal and Accountability}}. 
There are ethical issues for some applications of AI. For example, is it ethical to use AI to determine who should be paroled, or to predict who is most likely to commit a crime in the future? There are legal and regulatory issues regarding the fair use of AI in society. There are questions about the governance of AI systems and accountability for AI systems that need to be considered \cite{QIAN2024100040}.
\item{\textbf{Broad social consequences}}. 
Finally, there are a broad range of social concerns regarding AI, perhaps the strongest is the \emph{singularity}: will AI supercede human intelligence with profound consequences for human society and world power? Another important concern is the potential impact of AI on the labor markets and human employment (see e.g. \cite{dehouche2025post} on post-labor economics), and also on the environment due to excessive energy or water use by large AI systems. Furthermore, how will AI affect human relationships, both personal relations (with AI and amongst humans) and in wider society?
\end{description}
Trustworthy AI endeavours to mitigate and control these risks in AI systems.
Conformal Prediction (CP) is a technology to augment machine learning models by providing reliable measures of uncertainty \cite{Vovk2005AlgorithmicWorld}. 
In this article, we explore how their use can help address some of these risks and therefore support development of trustworthy AI.

\begin{figure} 
\centering
\includegraphics[scale=0.6] {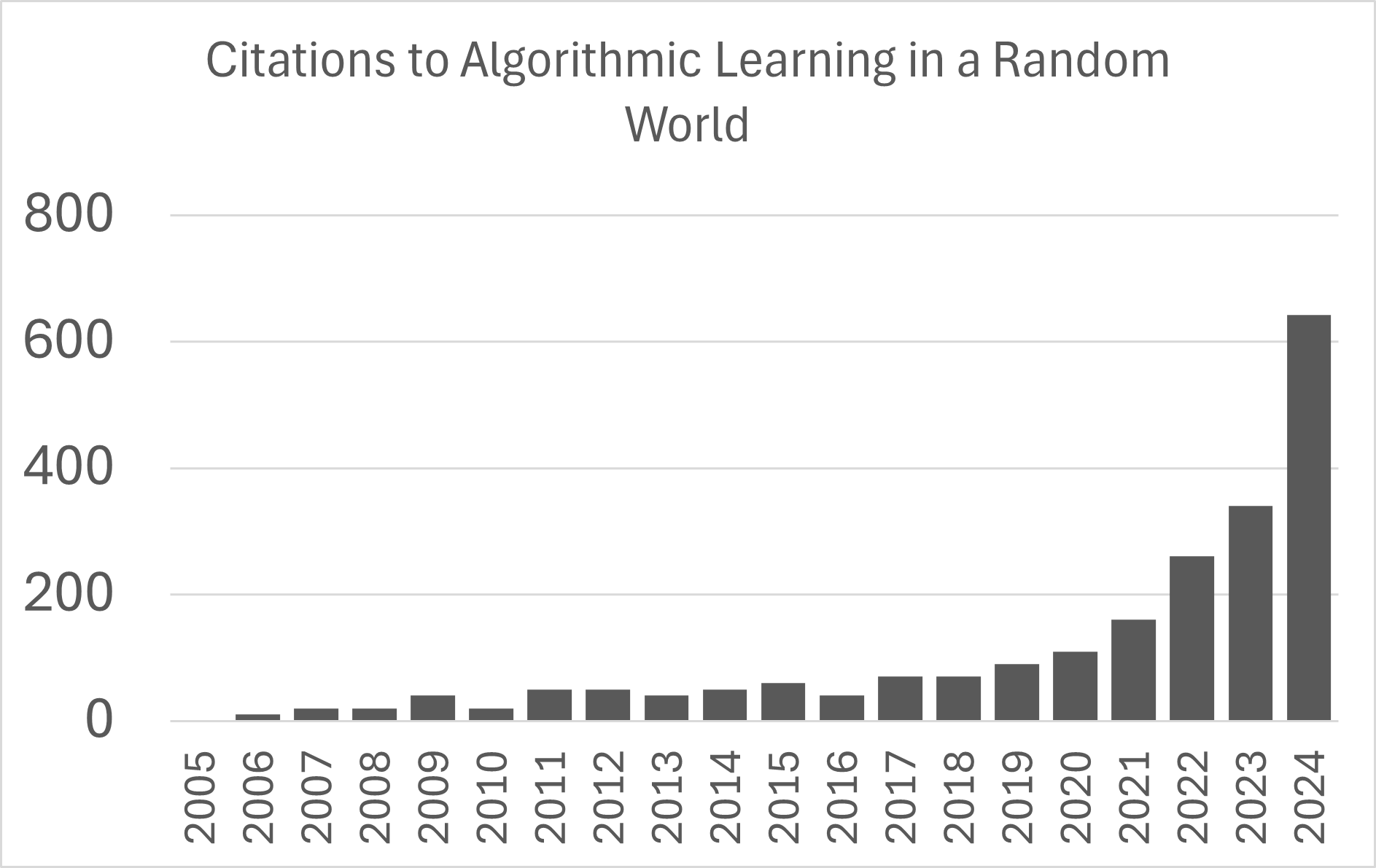}
 \caption{Annual citations to \cite{Vovk2005AlgorithmicWorld}, as measured by Google Scholar (accessed on 22 July 2025).}
\label{fig:alrw-citations}
\end{figure}

The CP framework was originally developed by Gammerman, Vovk and their research group in the 1990’s. One of their earliest publications, with Vapnik, introduced the exchangeability and predicting with confidence framework in the context of Support Vector Machines \cite{10.5555/2074094.2074112}. Later the algorithms reached maturity with the important validity results when they were first named confidence machines \cite{Papadopoulos2002InductiveRegression,vovk2002line}. Since then, with the need to develop reliable and trustworthy AI, interest in CPs has grown considerably, especially over the past few years, as is evident in Figure \ref{fig:alrw-citations} which shows citation rates for the key reference book in the field \cite{Vovk2005AlgorithmicWorld}.

AI is data-driven and most AI tools use deep neural networks which are trained on large datasets. 
As such, often the source of many AI risks can be found with problems in the data, and hence it makes sense for solutions to be found at the data collection and pre-processing steps of AI development \cite{hand2020dark}. 
Nevertheless, it is also possible to find solutions that can be implemented in the machine learning algorithms themselves.
Of the risks listed above the first three are amenable to technical solutions and hence these are the ones we will focus on in this paper. 
The fourth one, objective misalignment, can be mitigated by better management of the process of task specification, and so align with better software engineering processes and validation in machine learning.
The fifth one, security and privacy, is also amenable to a technical solution, but in the domain of computer security, rather than by applying machine learning algorithms such as CP.
Interestingly, the sixth risk for legal and regulatory responsibility, although a human, management and governance issue, may be informed by use of CP, as we will discuss later.

Predictive machine learning algorithms work by \emph{learning} a function that takes an input example and outputs some prediction. This may not be sufficient in many domains since the end-user may want to know how confident the algorithm is. For example, if a medical diagnosis system suggests that a patient has a certain subtype of leukaemia, it would be valuable if the doctor also knew the probability that this is the case. Many machine learning algorithms are able to do this by providing probabilities for the prediction. These probability estimates are typically based on underlying distributional assumptions. With a frequentist approach, an implicit assumption of normality is often required; whereas, for a Bayesian approach, an explicit assumption of a prior distribution is required \cite{Vovk2005AlgorithmicWorld}. 
CP works in a different way by providing a range of possible predicted values as a prediction set at a \emph{guaranteed} level of confidence. 
This guarantee is achieved as a mathematical property, with exchangeability being the only distributional assumption on the data. This is a minimal assumption which is even weaker than the usual independently and identically distributed (i.i.d.) assumption, which has led scholars to refer to CP as \emph{distribution-free} \cite{lei2014distribution}.
This is a valuable property, especially for trustworthy machine learning, as we will see. 
%Example \ref{example1} provides an illustration of how CP can be used for reliable uncertainty estimation in machine learning.

In Section \ref{sec:cp} we give a brief overview of CP, and then in Section \ref{sec:cp-airisks} describe how CP can help develop trustworthy AI systems.

\section{Conformal Predictors} \label{sec:cp}

Conformal Predictors are set predictors. Instead of predicting one possible value, called a \emph{point prediction}, they can predict a set of possible values for a target outcome. The set prediction is also referred to as a \emph{region prediction} in the literature.
A prediction set is correct if the true target value is in the prediction set, and wrong otherwise. Example \ref{example1} illustrates this approach.
\begin{framed}
\begin{example} \label{example1} 
Suppose that the machine learning task is to model a particular disease group. 
For a given patient, it is required to predict whether they have disease A, B, C or D. 
\begin{itemize}
\item A basic machine learning algorithm, given the details and symptoms of a new patient John Smith, may predict that he has disease B. But this provides no information on how confident we should be in this point prediction.
\item A probabilistic machine learning algorithm may be able to predict disease B with probability 75\%. This is more useful but depends on whether the assigned probability is well-calibrated, and if the underlying assumptions of the probabilistic model are correct. 
%It also does not provide alternative options for a medical doctor who may not be satisfied with the 75\% probability.
\item Finally, a Conformal Predictor produces the prediction set \{B, D\} at 95\% confidence level which means that the disease is either B or D with 95\% probability. 
This is valuable for the doctor since it means that she can take action to perform further tests to isolate which of these two it is, and the probability is reliable since CP has a guarantee of validity. 
The confidence level itself can be set by the doctors according to their own threshold. However, there is no ``free lunch'' and as confidence level increases, so the size of the prediction set tends to increase.
\item If later it emerges that John Smith actually has disease B, then the CP was correct. On the other hand, if it is disease C then the prediction set was wrong.
\end{itemize}
\end{example}
\end{framed}
The size of the prediction set is governed by a user-defined confidence level. 
A high confidence level can be set, which means we can be more certain of the prediction, but this can only be achieved with larger prediction sets: the CP will hedge its prediction by offering more options.
In the extreme case, setting 100\% confidence level means the prediction set will contain all possible target values, which is not any use.
This demonstrates an interesting quality of the CP: in other machine learning algorithms, performance will normally be measured by its accuracy, but in CP, this is guaranteed; instead, performance is related to the size of the prediction set: the smaller the better (except for the empty set which is a special case). This is called the \emph{predictive inefficiency} of the CP and there are several ways to measure this, although the mean set size (\emph{N} criterion) is typical \cite{Vovk2016}.

Since this article is discursive in nature, we will not dwell on the technical details. However, there is a rich mathematical foundation to CPs which the reader is invited to follow for details \cite{Vovk2005AlgorithmicWorld}.
However, we can give a brief overview of how CP works. At the heart of CP is a \emph{conformity measure} which is a function that takes an example (observation): both the input values and the target outcome and outputs how typical that example is. For example, if input is type of animal and the target variable is height in meters, then (``horse'', 2) would receive a high conformity score (say, 10) since 2 meters is a normal height for a horse, whereas (``mouse'', $5$) would be low (conformity score=$-3$) since a 5 meter mouse is very unlikely. 
The conformity measure is typically based on a machine learning model that has learned the typical relationship between the input and the target output.
The Conformal Predictor is an algorithm which takes the conformity measure to construct prediction sets for new unlabelled examples in such a way that the probability that the prediction set is correct is equal to a user-defined confidence level: this is the property of \emph{validity} 
\footnote{Technically, the probability of being correct is greater or equal to the confidence level, but so long as conformity scores are mostly unique, this becomes equality. It is easy to enforce this by adding small random noise to the conformity measure.}.
Remarkably, validity holds regardless of choice of conformity measure, so long as the data is exchangeable. However, poor choice of conformity measure will lead to high predictive inefficiency.
%Another way to think of the CP is that it takes a heuristic uncertainty measure, the confomity measure, and converts it to a reliable uncertainty estimate.

\begin{figure} 
\centering
\includegraphics[scale=0.6] {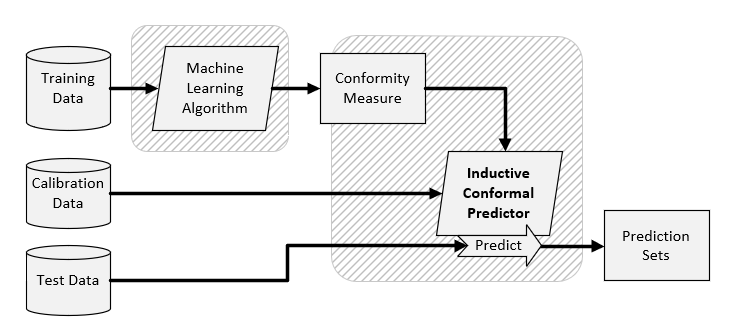}
 \caption{Inductive Conformal Predictor (ICP) (adapted from \cite{lei2023reliable}).}
\label{fig:icp}
\end{figure}

There are two types of CP: transductive CP (TCP) which operates in the online learning setting, and the inductive CP (ICP) which operates in the offline, batch setting. Since the batch setting, with a fixed training dataset and separate test dataset, is the usual setting in practice, we will focus on ICP. For ICP, a separate calibration dataset is required to achieve validity. Figure \ref{fig:icp} provides a schematic for ICP.

CP can be used for both classification and regression learning tasks. Classification involves tasks where the target variable is a finite set of labels, just two for binary classification, and so the prediction set is also finite. However, for regression the target variable is a real number. This means the prediction set is a subset of the real numbers and is typically infinite. It is usual to treat the special case when the prediction set is a prediction interval since this is manageable. The normalized conformity measure proposed by \cite{Papadopoulos2002InductiveRegression}, or some variant, is typically used for regression and yields prediction intervals.

Traditionally Conformal Predictors are \emph{wrapper} algorithms, in the sense that a machine learning model is first built and then it becomes to the input to a conformity measure which is then used by the Conformal Predictor to produce reliable prediction sets, as illustrated in Figure \ref{fig:icp}. Therefore, all the learning is done by the machine learning algorithm and CP provides reliable prediction sets.
The problem with this is that the learning is not optimized to the CP task. A new approach is to integrate the Conformal Predictor into the optimization and learning process. Studies have shown that this can lead to training Conformal Predictors which have lower predictive inefficiency for both classification \cite{stutz2022} and regression tasks \cite{lei2023reliable}.

\section{CP for AI Risk Mitigation} \label{sec:cp-airisks}

There are several aspects of AI Risk that CP can help reduce or avoid. In this section, we break down the risks in detail and describe how CP can contribute.
Typically, all machine learning models are evaluated according to aggregate performance on a task. If a classification task is learnt, then performance can be measured by accuracy, area under the ROC curve (AUC), F1-score and a myriad of other measures. For a regression task, typically mean-square-error, $R^2$ or mean-absolute-error. Academic articles reporting machine learning results will provide such aggregate performance measures and industry projects will also report these performance measures so, as long as the evaluation has been performed correctly, this is not usually a risk \cite{provost2013data}. However, there are several AI risks that may not typically be addressed and so deserve special attention.

\subsection{Performance - Calibration Risk}
When we use a machine learning algorithm to make a prediction, we would like to know how reliable this prediction is. This is different from accuracy. For example, if I consult a weather forecasting system, if it says that there is 75\% chance of rain I need to trust the probability really is 75\% (and not 65\% or 85\%). This is called \emph{calibration} and if a model's probability estimates are measurably evident in operation we say it is \emph{well-calibrated}. Then we can trust its uncertainty estimates are reliable.
Calibration can be tested over the long run by observing its prediction against outcome. For example, if the weather forecasting system makes 50 daily forecasts of rain between 70\% and 80\% over the course of the year, and of those days, 38 had rain, that would be about right. If it was only 20 (40\% empirical frequency) we would not trust it is well-calibrated. Statistical tests can be used to test for calibration; e.g. using a binomial test.

Conformal Predictor is a set predictor that outputs set predictions at a given confidence level. In this context, we say that a set predictor is well-calibrated if the frequency that the set predictions are correct is the same as the confidence level. As we have explained, CPs have the important property of validity which means that they are guaranteed to be well-calibrated, and this is true in the finite sample setting. This remarkable feature clearly ensures they are trustworthy for calibration, which means they are valuable tools when applying machine learning in safety-critical applications. 
Of course, the validity guarantee is dependent on both the data being exchangeable and a lack of coding errors when developing the prediction software (e.g. processing the data wrongly). For this reason it remains important to test the empirical validity.

\begin{figure} 
\centering
\includegraphics[scale=0.6] {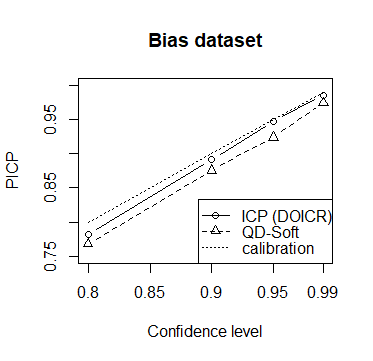}
\includegraphics[scale=0.6] {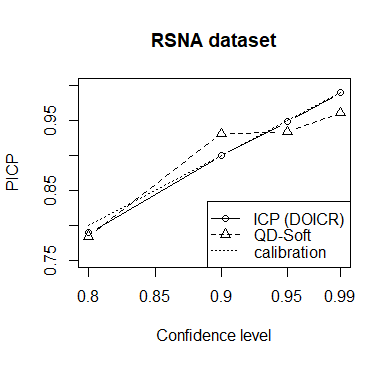}
 \caption{Experimental results showing calibration plots of ICP and QD-Soft algorithms, from \cite{lei2023reliable}, Tables 4 \& 5.
The two experiments were building neural networks on the bias correction data set (left) and building convolutional neural network (EfficientNet) on RSNA imaging data for bone age prediction (right). 
DOICR is  the authors directly trained ICP for regression, and PICP is prediction interval coverage percentage.
The calibration line is the target when PICP is equal to confidence level (perfectly calibrated).
} 
\label{fig:calibration}
\end{figure}

Not all set predictors are well-calibrated. One study performed experiments with several prediction interval algorithms, essentially set predictors in the regression setting \cite{lei2023reliable}. They found that whilst ICP was evidently well-calibrated, an alternative algorithm QD-Soft \cite{Pearce2018} was not. Figure \ref{fig:calibration} shows results for the bias correction \cite{cho2020comparative} and RSNA data sets \cite{halabi2019rsna}. The left graph shows that both models' coverage is slightly lower than the confidence level; however, ICP is superior being closer to the calibration line, especially at higher confidence level. The right graph shows a clear difference with ICP well-calibrated, whilst QD-Soft is extremely over- and under-calibrated, especially at high confidence levels.

The validity property is the key reliability property that CPs have, hence when they are used for reliable or trustworthy machine learning, it is usually the calibration risk that is addressed.
For example, CP is used to develop trustworthy deep learning in the clinical setting \cite{gamble2024toward}.
In \cite{lu2022improving}, disease severity rating based on medical imaging is made reliable using CP. They provide experimental results to demonstrate the successful use of conformal predictors in this setting.
In another study, CP is integrated into a multi-view (multi-model) deep learning framework, using the Conformalized Multi-view Deep Classification, to enable prediction set prediction with validity for trustworthy models, and provide results for medical imaging data \cite{liu2024building}.
However, calibration is not the only AI risk that can be addressed by CP and the next sections will address these issues.

\subsection{Performance - Generalization Risk} \label{sec:genrisk}
Although the model may be built and perform well on the development data, it may not generalize well to new data. This is important because the purpose of a model is to predict on new, previously unseen, data. There are different forms of this risk.
The problem of models \emph{overfitting} to training data is well-known. If a model overfits, it will not generalize well beyond the training data. However, this problem is largely dealt with by setting up a proper validation procedure, ensuring that the model is evaluated on data that is independent of data that it was trained on.  Randomly splitting development data into training and test data is the easiest approach to do this, or cross-validation is an alternative.

More challenging versions of generalization risk are due to (i) Selection bias: the population in development data is different in some way to the data that the model will be applied to; (ii) Population drift: the time period of the development data is different from the data the model will be applied to in the future. 
An example of selection bias is when a credit score model is built on a historic portfolio of credit card holders who are on average old and predominantly male, but the model will be applied on a population of applicants who are expected to be younger with a higher proportion of females. Such a model may not perform so well on the target population because of difference in behaviour across age groups and sex.
In the medical context, an example of population drift may occur with the development of CT scanner imaging technology over time. A machine learning algorithm built on old CT images may not work well on future ones. Other sources of drift could be change in patient demographics, change in disease prevalence and behaviour, change in disease definition and understanding, or change in physician behaviour, perhaps as a result of automation \cite{10.1259bjr.20220878}.
What all forms of generalization risk have in common is that there is a distributional change between development and target operational data. 

We can utilize the property of CP that if data is drawn from the same exchangeable distribution then a CP built on that data will exhibit validity.
By \emph{modus tollens}, this means that if we detect that a Conformal Predictor is not exhibiting empirical validity then the data is not exchangeable and not drawn from a single homogeneous distribution. This would signal a potential generalization risk.
In this way, CP can be used as a generalization risk warning system and this is the underlying idea of \emph{conformal testing}.
 A framework can be set up using conformal test martingales based on the p-values output by the CP making sequential predictions, as described in \cite{pmlr-v152-vovk21b}. Although the paper focusses on drift, it could be adapted for the selection bias setting. 
 A more recent development of this idea is conformal e-testing for distribution change \cite{VOVK2025111841}.
 There are at least two ways this conformal testing methodology can be used to reduce generalization risk.
\begin{itemize}
\item Conformal testing is a warning system to determine when to rebuild a model. 
\item There are several methods proposed to reduce the impact of generalization risk such as resampling or weighting the development data. These often involve approaches to make development data similar to target data in some sense \cite{bellotti2020b}. Conformal testing can be used as a way to test whether these approaches have worked successfully, or not.
\end{itemize}

\subsection{Performance - Robustness Risk}

An AI system is robust if it is able to operate well even under unusual conditions; in particular, if it is given a new example unlike those it has been trained on, we would still like it to perform well. Unfortunately, many AI systems lack this type of robustness. If they are given data that is quite different, or untypical, of training data, they behave poorly (see e.g. \cite{9798161}). This is related to generalization risk, but these cases are typically rarer and may lead to more extreme outcome.
A real example is illustrated by the case of Tesla cars operating in autopilot on the highway with 11 accidents between 2018 and 2021. The accidents all had the same common feature: the car crashed into a stationary emergency vehicle. The accidents were investigated by the National Highway Transportation Safety Administration (USA). This particular scenario, a stationary emergency vehicle on the highway, was unusual for the car's autopilot, since it would not have emerged during training. Unfortunately the autopilot was not able to adapt to the scenario in a way we would expect a human driver to do \cite{isadore2021}.
Sometimes a very small change can cause a machine learning model to fail; e.g. \cite{8601309} showed that changing just one pixel in an image can upset a deep neural network. This vulnerability to very small changes is referred to as \emph{brittleness}.

CP can be used to avoid problems when new examples are different from the training data. This takes advantage of the conformity measure in the CP. The conformity measure scores how typical a new example is, therefore an unusual case - one not seen in training - will yield a low conformity score. It would be expected that such a case would yield a very inefficient prediction set, since it is untypical, which would be a way for the AI system to signal that it is not confidently predicting in these cases. 
The conformity measure can be designed especially to find unusual cases not just in the relationship between the input features and the target variable but also amongst the input features themselves.
In particular, \cite{laxhammar2015inductive} proposes a nearest-neighbors-based nonconformity measure that can be used for this purpose. In their experiments using a dataset of real sea vessel trajectories, they were able to achieve high accuracy levels for identification of anomalous trajectories with low false alarm rates. Furthermore, the anomaly detection is shown to be well-calibrated.

Therefore, by using CP to detect anomalous examples, it is possible to provide a warning regarding unusual examples and situations that the AI system can act upon to ensure safety.

\subsection{Bias  - Biassed Prediction Risk}

Bias has a number of different meanings. In the context of trustworthy AI, we usually refer to decisions made by AI systems that are biassed towards some particular group of people. For example, it could be bias towards women, or black people, or Jewish or Muslim people. We would say that the AI system discriminates against those people. In many countries, there are protected characteristics by law. For example, in the UK, the Equality Act 2010 protects people in law based on age, disability, sex, sexual orientation, gender reassignment, race and several other categories. Therefore, AI systems that are biassed may break the law. 
The bias may be more specific, and hence more difficult to uncover, crossing two or more protected characteristics: such as discrimination toward black women \cite{buolamwini2018gender}. This is known as intersectionalism.

Bias and discrimination are linked to notions of unfairness in algorithmic decision making. Fairness is well-understood and has been defined clearly in the context of machine learning \cite{barocas-hardt-narayanan}.
One key insight is that fairness is not a single idea, but can be defined in many ways. Common definitions of fairness are Independence, Separation, and Sufficiency. These definitions are contradictory, in the sense that if we choose one type of fairness, we cannot achieve the others \cite{barocas-hardt-narayanan}.
In practice, this means that the stakeholders of an AI system need to agree on the form of fairness that is required for the application, and this will be strongly related to the legal and regulatory framework.
This decision will then affect how we measure whether our model is unfair.
%When discussing CP solutions below, it is worth considering that they could be adapted to the various fairness definitions that are available.

It is worth pointing out that not all bias may be unethical or unfair. For example, a medical system may find that men are more likely than women to contract a disease, all else being equal. However, this may reflect an actual bias in reality, i.e. men are genuinely more susceptible to the disease, rather than representing a poor diagnosis in women. Therefore, a judgement needs to be made regarding whether a biassed outcome represents unfair discrimination.

Bias is well-known in AI and has been well-documented. The possible causes of bias are myriad:  
bias in the way data is selected, historical bias in the data, bias introduced (or amplified) by the algorithm, bias by the machine learning engineer. If the bias is detected and the cause of the bias is isolated then a solution to mitigate the bias may be proposed \cite{app14198860}.

Predictions may be biassed within subgroup. So, overall a model may predict accurately across a whole population but for some subgroups it may be that the target variable is over- or under-predicted. For example, in 2015 Amazon ditched a project to use AI in recruitment, since the model was under-scoring women applicants, specifically penalizing applicants with female-linked descriptions in their application (e.g. ``women'' or ``women's chess club captain'') \cite{dastin2022amazon}.
This type of bias could be fixed by either returning to the data to adjust for any under-representation, or historical bias, or, in the context of supervised learning adjusting for failures in the model by building a second level model of the errors, to adjust post-hoc to the bias. Interestingly, machine learning algorithms based on boosting can do this by design \cite{app14198860}.

\subsection{Bias - Biassed Performance Risk}
\label{sec:bias-perf-risk}

Another form of bias is not with the prediction itself, but with the performance of the model for different subgroups. 
We have already given the example of intersectional bias for facial recognition systems \cite{buolamwini2018gender} in the Introduction.
In another study looking at diagnosis based on medical imaging data and using machine learning found that imbalanced data sets lead to models that reduce performance in diagnosis for the underrepresented gender, measured using AUC \cite{larrazabal2020gender}.

For CP, the key performance measure is predictive inefficiency, and so biassed performance could manifest in changes in prediction set size between subgroups. For example, if we observe larger prediction sets for an ethnic minority, on average, then this would suggest bias that will require further investigation. Hence, CP developers would be recommended to check for this. Mitigating this will require similar techniques as other machine learning algorithms (e.g. analysis and adjustment of data). However, as mentioned in Section \ref{sec:genrisk}, prediction inefficiency is measured without knowing outcome, so CP for bias performance tests will be particularly important in applications with delayed outcome, such as patient survival or credit scoring, where actual outcome may not be known for several years.

Another way that CP may exhibit biassed performance is with its coverage: the proportion of times that a prediction set is correct within subgroups. This is because the validity property referred to in Section \ref{sec:cp} is across the entire population, but may not be true for all subgroups, as illustrated by Example \ref{example2}. This means the validity property is \emph{marginal validity}, but to ensure even coverage between groups, we are interested in is \emph{conditional validity}; i.e. validity conditional on the subgroups.
Arguably, conditional validity is more important for CP, rather than biassed prediction set size, since the validity property is the main benefit of CP over other algorithms.
Unfortunately, it has been shown that conditional validity is impossible in the general finite-sample case \cite{vovk2013}.
Approaches have been taken for special cases, such as linear models \cite{McCullagh2009} or asymptotic case \cite{lei2014distribution}, but these cannot be extended to the general case.
In terms of fairness, conditional validity fits the Independence criterion \cite{barocas-hardt-narayanan}, since validity, meeting the confidence level, across all observations is the central benefit for CP. 
We could imagine situations when we might want to achieve different coverage conditional on the outcome which would lead to the Separation definition, but these are not intuitive and would be special cases.

\begin{framed}
\begin{example} \label{example2} 
Suppose the CP task is to model a particular disease group, and for a given patient, predict whether they have disease A, B, C or D. The medical doctor sets a confidence level of 90\%.
\begin{itemize}
\item We observe that in practice, the CP gives empirical coverage of 89.5\%. This means that the prediction sets contain the true label (for each patient) 89.5\% of the time. 
This is sufficiently close to 90\% that we can conclude that the CP is well-calibrated.
\item If we consider just the male patients, we find that the empirical coverage is higher at 94\%. \item Conversely, the female patient group has 85\% coverage.
\item If the male/female populations are the same size, then this shows that overall empirical coverage is consistent with 89.5\%. 
\item However, it is clear that the CP is not well-calibrated \emph{within} these subgroups, and is under-performing for women.
\end{itemize}
\end{example}
\end{framed}

Mondrian CP \cite{Vovk2005AlgorithmicWorld} has also been proposed and gives conditional validity within the context of a given taxonomy of subgroups. Essentially, these work by splitting the calibration data into separate calibration sets for each subgroup. This guarantees conditional validity within these subgroups. However, Mondrian CP has two problems: (i) As taxonomy gets larger, the calibration sets get smaller, hence potentially degrading their performance; (ii) Mondrian is dependent on the pre-defined taxonomy and cannot guarantee validity in other subgroups.

Another approach is to aim for approximate conditional validity instead of exact.
In \cite{Barber2020}, a \emph{distribution-free approximate conditional coverage} framework is proposed that takes conditional validity to be true for at least some pre-defined proportion of the population. This is similar to the PAC-type conditional validity proposed by \cite{Vovk2014}. In particular they further propose \emph{restricted conditional coverage} to provide local conditional validity in the neighbourhood (ball) around some point. This is a promising approach and they show that CP satisfies this condition, in a special case, but they identify computational difficulties that require further research.
Another proposal for approximate conditional validity is the \emph{iterative feedback-adjusted conformity measure} (IFACM) algorithm that uses a boosting approach to update the conformity measure based on the performance of a conformal predictor in different subgroup regions \cite{bellotti2021approximation}. It can be set up with a taxonomy like Mondrian, but also has the advantage of being general-purpose, discovering subgroup regions with performance bias itself, without the need of a taxonomy. 

Although conditional validity is a useful goal, and Mondrian and approximation approaches are promising tools, it is clear that these approaches work by shifting predictive inefficiency to improve coverage, and there is ``no free lunch''. In particular, IFACM works explicitly by expanding or contracting prediction set sizes in response to deviations in validity across the population.
As such, although they achieve approximate conditional validity, this will typically translate into changes in predictive inefficiency, often increasing. 
This is acceptable since the main goal of CP is to achieve reliability. However, since many bias issues are often data-driven \cite{hand2020dark}, e.g. underrepresented sub-populations, it will still be valuable to fix the problem at the data and pre-processing level.

\subsubsection{Experiments using Mondrian and IFACM for performance bias mitigation}

We implemented experiments using Mondrian and IFACM on two real datasets, ASCIncome dataset \cite{ding2021retiring} and ISIC2019 dataset \cite{tschandl2018ham10000,codella2018skin,hernandez2024bcn20000}. We selected three subgroups which are detailed in Table \ref{tab:subgroup} for each dataset. Subgroup 1 is a majority group, subgroup 2 and subgroup 3 are minority groups. 

\begin{figure}[ht]
    \centering
    \begin{subfigure}[b]{0.8\textwidth}
        \includegraphics[width=1.0\linewidth]{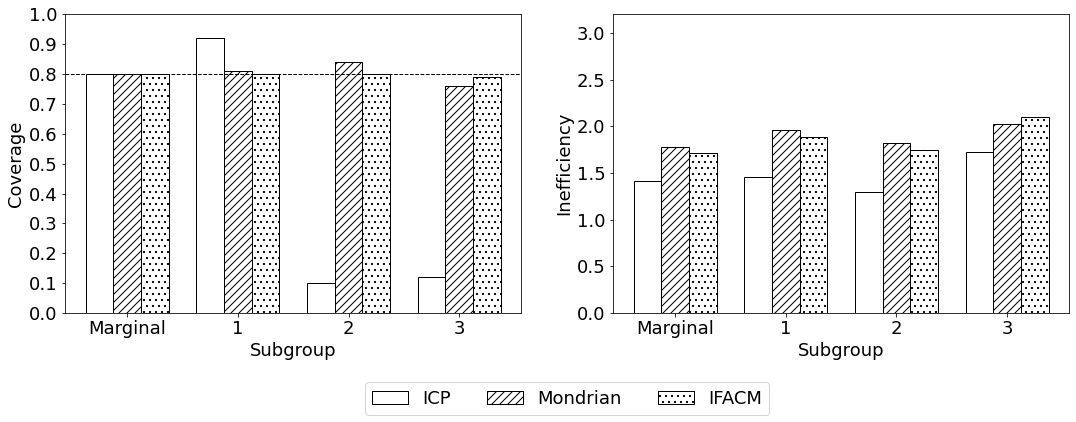}
        \caption{Confidence level=0.8 \\[4ex]}
        \label{subfig:acs_8}
    \end{subfigure}
    \begin{subfigure}[b]{0.8\textwidth}
        \includegraphics[width=1.0\linewidth]{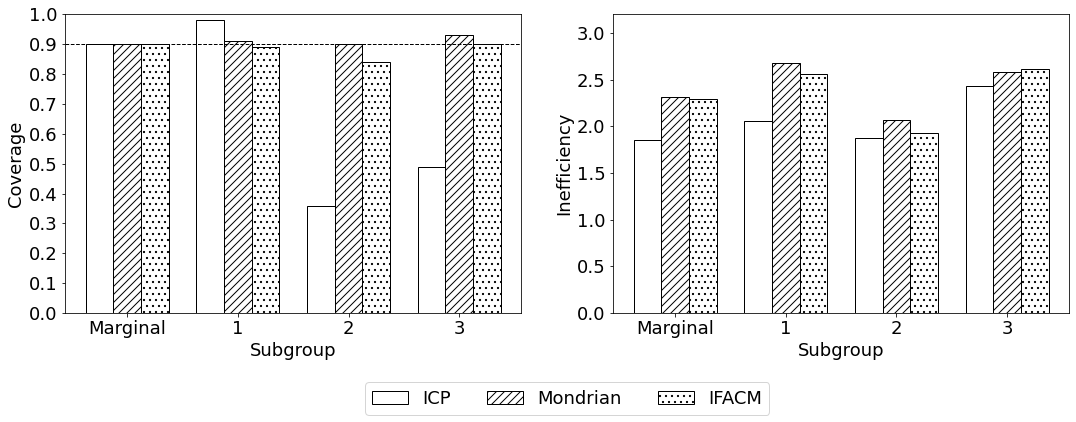}
        \caption{Confidence level=0.9}
        \label{subfig:acs_9}
    \end{subfigure}
    \caption{Experimental results on ASCIncome dataset for performance bias correction with (a) confidence level$=$0.8 and (b) confidence level$=$0.9.}
    \label{fig:acs}
\end{figure}

In the ASCIncome experiment, we built a neural network model (i.e. a Multi-Layer Perceptron with 64, 32, 32 and 8 hidden units in each layer) to classify annual income levels based on individual demographic features including age, educational attainment, occupation, gender and race. We perform an expository analysis of the results to illustrate performance bias correction. Figure \ref{fig:acs} shows the empirical coverage and inefficiency within the subgroups obtained by the basic ICP, Mondrian and IFACM, with confidence levels 0.8 and 0.9. We see that all the methods achieve marginal validity. However, ICP exhibits coverage higher than the confidence level conditional on subgroup 1 and extremely low coverage conditional on subgroup 2 and subgroup 3. By contrast, Mondrian and IFACM provide coverages that are very close to the confidence level across all subgroups, albeit at the cost of increasing inefficiency. These results demonstrate the significance for applications where policymakers express major concern about the validity conditional on age, gender and race, and do not have any preference for other demographic features such as educational attainment and occupation.

\begin{figure}[ht]
    \centering
    \begin{subfigure}[b]{0.8\textwidth}
        \includegraphics[width=1.0\linewidth]{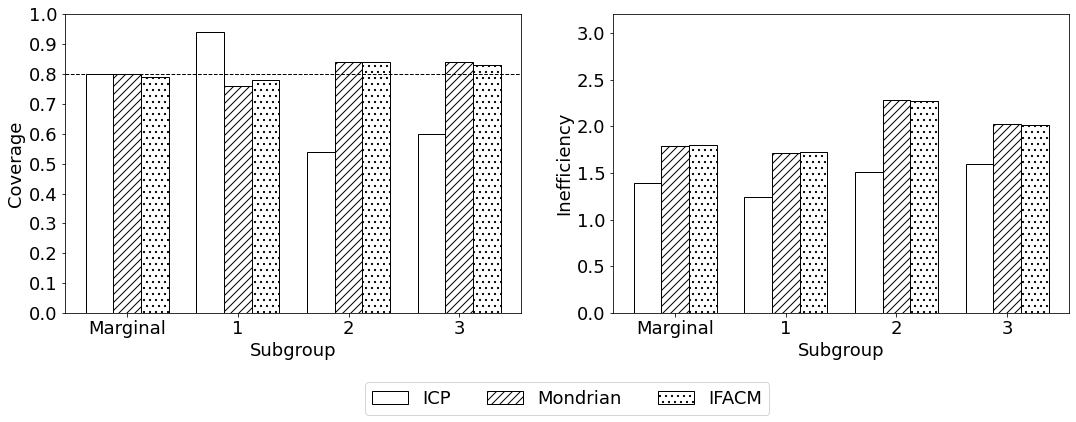}
        \caption{Confidence level=0.8 \\[4ex]}
        \label{subfig:isic_8}
    \end{subfigure}
    \begin{subfigure}[b]{0.8\textwidth}
        \includegraphics[width=1.0\linewidth]{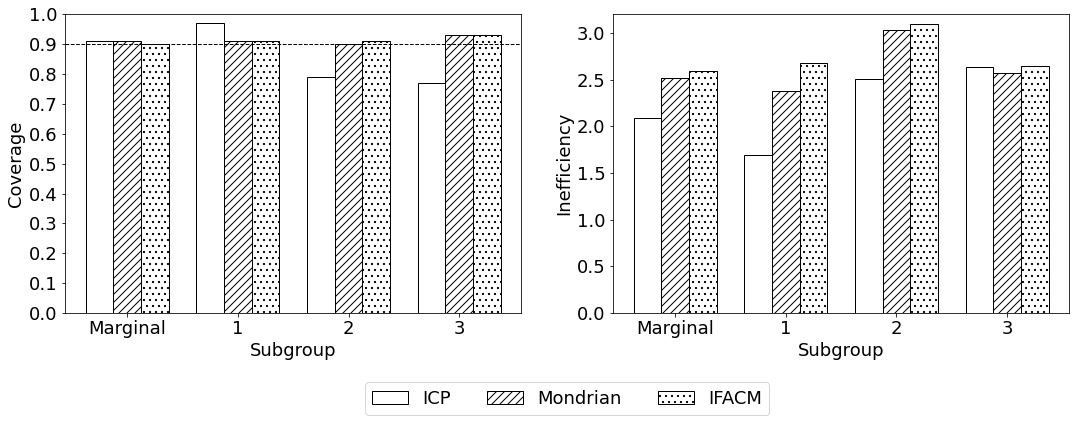}
        \caption{Confidence level=0.9}
        \label{subfig:isic_9}
    \end{subfigure}
    \caption{Experimental results on ISIC2019 dataset for performance bias correction with (a) confidence level$=$0.8 and (b) confidence level$=$0.9.}
    \label{fig:isic}
\end{figure}

In the ISIC2019 task, we built a ResNet50 model to classify diagnostic categories given the dermoscopic images of the skin lesion. Figure \ref{fig:isic} shows the empirical coverage and inefficiency within the subgroups obtained by ICP, Mondrian and IFACM, with confidence levels 0.8 and 0.9. All of the methods achieve marginal coverages that are close to the confidence level. ICP achieves higher coverage conditional on subgroup 1 and significantly lower coverage conditional on subgroup 2 and subgroup 3. Mondrian and IFACM provide conditional coverages that are closer to the confidence level across all subgroups, albeit with higher inefficiency. The results demonstrate that Mondrian and IFACM are potential approaches to increasing trustworthiness in medical AI by providing validity across clinically relevant subgroups.

\begin{table}[htbp]
\centering
\begin{tabular}{|l|l|l|}
\hline
\textbf{Subgroup} & \textbf{ASCIncome} & \textbf{ISIC2019} \\
\hline
1 & female, age$>$55, White, 2k$\leq$income$<$70k   & female, age$\leq$40, Melanocytic nevus   \\
\hline
2 & female, 30$<$age$\leq$55, Black, income$\geq$140k & female, 40$<$age$\leq$55, Benign keratosis \\
\hline
3 & male, 30$<$age$\leq$55, Asian, income$\geq$140k   & male, age$>$70, Benign keratosis           \\
\hline
\end{tabular}
\vspace*{0.5 cm}
\caption{Subgroup descriptions}
\label{tab:subgroup}
\end{table}

\subsection{Human/AI Reliance and Regulatory Risks}

When AI systems are deployed, the response of people using the system is sometimes not what we would expect or desire. Human-AI interaction is a special topic area of human-computer interaction, studying specifically the behaviour of people when working with AI, and ways we can design AI for better collaboration with humans (e.g. \cite{abedin2022designing,mehrotra2024systematic}). Much of the misuse or disuse of AI, which can cause failures such as railway crashes and biassed decision making comes from people over- or under-trusting the AI, and this is ``often linked to the alignment between the perceived and actual performance of the system'' \cite{mehrotra2024systematic}.
This is particularly the case with systems that do not provide a confidence measure for their predictions or confidence measures that are not well-calibrated. 
In these circumstances, the user has no way to measure the AI's trustworthiness and this can lead to poor outcomes.
CP can mitigate this problem since it is able to provide reliable uncertainty measures that can be trusted. 
In particular, if approaches that deal with conditional validity such as Mondrian and IFACM, illustrated in Section \ref{sec:bias-perf-risk}, are deployed, then human users will have additional confidence at subgroup and indivual decision-making level.

With the increased adoption of regulation and laws on AI, such as the European Union AI Act, there is an emphasis on safe, trustworthy and transparent AI \cite{eu-ai-reg}. The qualities of CP make it a valuable tool to demonstrate the reliability and transparency of uncertainty measures in AI decision making to regulators.

\section{Conclusion}

In this article, we have considered how CP can be used broadly for trustworthy AI, and given some specific experimental results to demonstrate the value of CP for calibrated uncertainty estimation and bias detection and mitigation.
Since many of the risks are data-driven, we acknowledge that many solutions to enable trustworthy AI fall back to careful data collection, processing and use \cite{hand2020dark}. Nevertheless, we show that CP can provide some valuable algorithmic tools.

Table \ref{tab:overview} provides an overview of AI risks, along with characteristics of CP that can be beneficial. Some of the AI risks have not been addressed, since they are not so obviously addressed by CP. Most previous articles that discuss CP in the context of trustworthy AI typically do so in the context of marginal validity and calibration (e.g. \cite{gamble2024toward,lu2022improving}) However, future work could expand the scope of CP for trustworthy AI further in areas identified in this paper, and perhaps also in novel areas of AI risk that are not addressed by CP yet according to Table \ref{tab:overview}.
\begin{table}
\begin{center}
\begin{tabular}{ |p{3.5cm}|p{3.5cm}|p{5cm}| }
\hline
AI risk category & AI risk & CP mitigation \\
\hline
Poor performance & Calibration risk & Validity guarantee \\
& Generalization risk & Conformal testing \\
& Robustness & Conformity measure / anomaly detection \\
\hline
Bias & Biassed predictions & - \\
 & Biassed performance & Mondrian and approximate conditional validity (IFACM) \\
 \hline
Human-AI interaction & & Reliable prediction sets \\
\hline
Objective misalignment & & - \\
\hline
Security \& Privacy & & - \\
\hline
Ethics, Legal \& Accountability & & Reliable uncertainty measures for reporting, accountability and transparency \\
\hline
\end{tabular}
\end{center}
\caption{Summary of AI risks and CP mitigation} \label{tab:overview}
\end{table}

Although this article focusses on predictive machine learning, new research is emerging applying CP to generative models such as large language models (LLMs) \cite{wang2024conuconformaluncertaintylarge,10.1162/tacl_a_00715}. Consequently, we can foresee CP also being used for trustworthy and reliable AI in this expanding field of AI with future research.
Finally, we note the value of CP to allow for better understanding and interpretation of AI, and therefore an important tool for AI governance, to demonstrate adherence to standards and regulations as these emerge throughout the world both in industry and government.

\newpage % force new page for formatting
\section*{Acknowledgment}
We are grateful for fruitful discussions on trustworthy AI over many years with the team at Validate AI \footnote{\texttt{https://validateai.org} (accessed on 25 July 2025)}, especially Shakeel Khan, David Hand and Mark Kennedy.
We are especially indebted to Alex Gammerman who introduced Anthony Bellotti to CP in the early 2000's and enabled him to complete his PhD using CP in 2006. Xindi Zhao is subsequently Anthony's PhD student and will graduate with her PhD using CP this year. Many thanks to Alex, and congratulations on his 80th birthday!

\bibliographystyle{splncs04}
\bibliography{CP-trustworthy-AI}

\end{document}